\documentclass[10pt,twocolumn,letterpaper]{article}

\usepackage{cvpr}              

\usepackage{graphicx}
\usepackage{float}
\usepackage{amsmath}
\usepackage{amssymb}
\usepackage{booktabs}
\usepackage{enumerate}
\usepackage{colortbl}
\usepackage[accsupp]{axessibility}

\makeatletter
\newcommand{\printfnsymbol}[1]{%
  \textsuperscript{\@fnsymbol{#1}}%
}
\makeatother

\definecolor{mygray}{gray}{.9}

\usepackage[pagebackref,breaklinks,colorlinks]{hyperref}

\makeatletter
\renewcommand{\maketag@@@}[1]{\hbox{\m@th\normalsize\normalfont#1}}%
\makeatother

\usepackage[capitalize]{cleveref}
\crefname{section}{Sec.}{Secs.}
\Crefname{section}{Section}{Sections}
\Crefname{table}{Table}{Tables}
\crefname{table}{Tab.}{Tabs.}

\def\cvprPaperID{571} 
\def\confName{CVPR}
\def\confYear{2022}

\begin{document}

\title{A Keypoint-based Global Association Network for Lane Detection}


\author{
Jinsheng Wang\textsuperscript{\rm 1,3}\thanks{Equal contribution} \quad Yinchao Ma\textsuperscript{\rm 2,3}\printfnsymbol{1} \quad Shaofei Huang\textsuperscript{\rm 3}\printfnsymbol{1} \quad Tianrui Hui\textsuperscript{\rm 4,5} \\ \quad Fei Wang\textsuperscript{\rm 2} 
\quad Chen Qian\textsuperscript{\rm3} \quad Tianzhu Zhang\textsuperscript{\rm 2}\thanks{Corresponding author (tzzhang@ustc.edu.cn)} \\
\textsuperscript{\rm 1} Peking University \quad \textsuperscript{\rm 2} University of Science and Technology of China \quad \textsuperscript{\rm 3} SenseTime Research  \\
\textsuperscript{\rm 4} Institute of Information Engineering, Chinese Academy of Sciences\\
\textsuperscript{\rm 5} School of Cyber Security, University of Chinese Academy of Sciences\\
}

\maketitle
\begin{abstract}
Lane detection is a challenging task that requires predicting complex topology shapes of lane lines and distinguishing different types of lanes simultaneously.
Earlier works follow a top-down roadmap to regress predefined anchors into various shapes of lane lines, which lacks enough flexibility to fit complex shapes of lanes due to the fixed anchor shapes.
Lately, some works propose to formulate lane detection as a keypoint estimation problem to describe the shapes of lane lines more flexibly and gradually group adjacent keypoints belonging to the same lane line in a point-by-point manner, which is inefficient and time-consuming during postprocessing.
In this paper, we propose a Global Association Network (GANet) to formulate the lane detection problem from a new perspective, where each keypoint is directly regressed to the starting point of the lane line instead of point-by-point extension.
Concretely, the association of keypoints to their belonged lane line is conducted by predicting their offsets to the corresponding starting points of lanes globally without dependence on each other, which could be done in parallel to greatly improve efficiency.
In addition, we further propose a Lane-aware Feature Aggregator (LFA), which adaptively captures the local correlations between adjacent keypoints to supplement local information to the global association.
Extensive experiments on two popular lane detection benchmarks show that our method outperforms previous methods with F1 score of 79.63\% on CULane and 97.71\% on Tusimple dataset with high FPS. The code will be released at \url{https://github.com/Wolfwjs/GANet}.
\end{abstract}

\begin{figure}[!t]
    \centering
    \subfloat[\label{fig:introa}]{
        \includegraphics[width=0.22\textwidth]{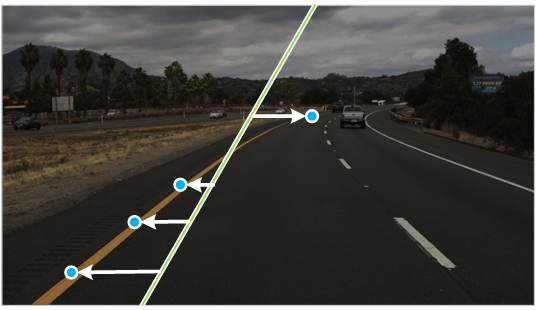}
    }
    \subfloat[\label{fig:introb}]{
        \includegraphics[width=0.22\textwidth]{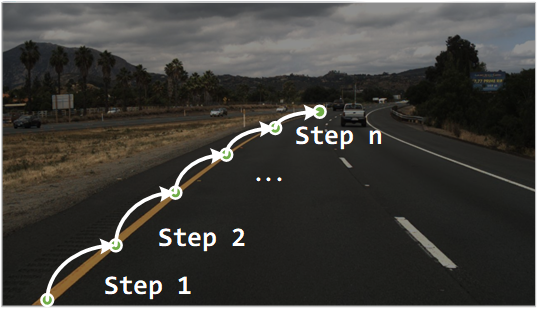}
    }\quad
    \subfloat[\label{fig:introc}]{
        \includegraphics[width=0.22\textwidth]{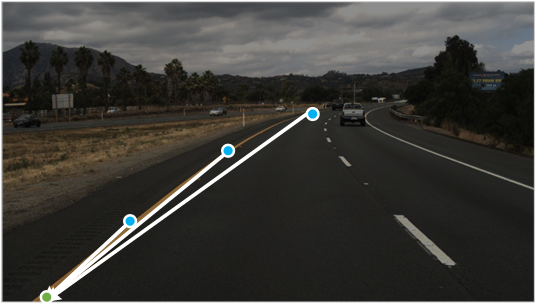}
    }
    \subfloat[\label{fig:introd}]{
        \includegraphics[width=0.22\textwidth]{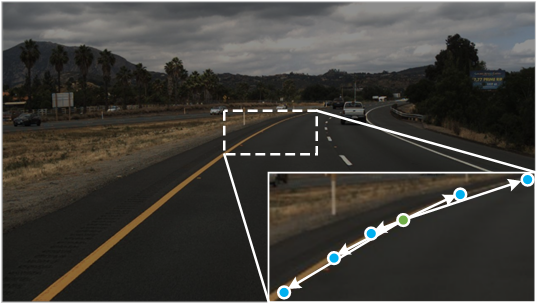}
    }
    \caption{(a) Anchor-based methods, which regress the predefined anchors into the shape of lanes. (b) Keypoint-based methods, which predict offsets between keypoint to its neighbourhood to group them one-by-one. (c) Illustration of our GANet, which directly regresses each keypoint to its belonged lane by predicting offset between each keypoint and the starting point of its corresponding lane line. (d) Illustration of our LFA module, which correlates each keypoint with its adjacent points for local information supplement.}
\end{figure}

\section{Introduction}
\label{sec:intro}
Autonomous driving~\cite{linecnn} has drawn remarkable attention of researchers from both academia and industry.
In order to ensure the safety of the car during driving, the autonomous system needs to keep the car moving along the lane lines on the road, requiring accurate perception of the lane lines.
Thus, lane detection plays an important role in the autonomous driving system, especially in Advanced Driver Assistance System (ADAS).

Given a front-viewed image taken by a camera mounted on the vehicle, lane detection aims to produce the accurate shape of each lane line on the road. 
Due to the slender shapes of lane lines and the need for instance-level discrimination, it is crucial to formulate lane detection task appropriately.
Inspired by the anchor-based object detection methods~\cite{ren2015faster}, some works~\cite{Tabelini_2021_CVPR,linecnn} follow a top-down design as illustrated in Figure~\ref{fig:introa}. 
Similar to object detection, a group of straight lines with various orientations are defined as anchors.
Points on anchors are regressed to lane lines by predicting the offsets between anchor points and lane points.
Afterward, Non-Maximum Suppression (NMS) is applied to select lane lines with the highest confidence.
Although this kind of method is efficient in lane discrimination, it is inflexible because of the predefined anchor shapes. 
The strong shape prior limits the ability of describing various lane shapes, resulting in sub-optimal performances of these methods.

To describe complex shapes of lane lines flexibly, Qu~\textit{et al}.~\cite{Qu_2021_CVPR} propose to formulate lane detection as a keypoint estimation and association problem, which takes a bottom-up design as illustrated in Figure~\ref{fig:introb}.
Concretely, lanes are represented with a group of ordered key points evenly sampled in a sparse manner. 
Each key point is associated with its neighbours by estimating the spatial offsets between them.
In this way, key points belonging to the same lane are integrated into a continuous curve iteratively.
Though keypoint-based methods are flexible on the shape of lane lines, it is inefficient and time-consuming to associate only one keypoint to its belonged lane line at each step.
Besides, the point-by-point extension of keypoints is easy to cause error accumulation due to the lack of global view.
Once a particular keypoint is wrongly associated, estimation of the rest part of the lane line will fail.

To overcome the above limitations, we formulate the lane detection problem from a new keypoint-based perspective where each keypoint is directly regressed to its belonged lane, based on which a novel pipeline named Global Association Network (GANet) is proposed.
As illustrated in Figure~\ref{fig:introc}, each lane line is represented uniquely with its starting point, which is easy to determine without ambiguity.
To associate a keypoint properly, we estimate the offset from the keypoint to its corresponding starting point.
Keypoints whose approximated starting points fall into the same neighborhood area will be assigned to the same lane line instance, thus separating keypoints into different groups. 
Different from previous keypoint-based method~\cite{Qu_2021_CVPR}, our assignment of keypoints to their belonged lanes is independent of each other and makes the parallel implementation feasible, which greatly improves the efficiency of post-processing.
Besides, the keypoint association is more robust to the accumulated single-point errors since each keypoint owns a global view.

Although keypoints belonging to the same lane line are integrated during post-processing, it is important to ensure the correlations between adjacent points in order to obtain a continuous curve.
To this end, we develop a local information aggregation module named Lane-aware Feature Aggregator (LFA) to enhance the correlations between adjacent keypoints.
To adapt to the slender and long shapes of lanes, we modify the sampling positions of the standard 2D deformable convolution~\cite{dai2017deformable} by predicting offsets to adjacent points to sample within a local area on the lane each time. 
In this way, features of each keypoint are aggregated with other adjacent points, thus acquiring more representative features.
We further add an auxiliary loss to facilitate estimating the offset predicted on each key point.
Our LFA module complements the global association process to enable both local and global views, which is essential for dense labeling tasks like lane detection.

Our contributions are summarized as follows:
\begin{itemize}
    \item We propose a novel Global Association Network (GANet) to formulate lane detection from a new keypoint-based perspective which directly regress each keypoint to its belonged lane. To the best of our knowledge, we are the first to regress keypoints in a global manner, which is more efficient than local regression.
    \item We develop a local information aggregation module named as Lane-aware Feature Aggregator (LFA) to enhance correlations among adjacent keypoints to supplement local information.
    \item Our proposed GANet achieves state-of-the-art performances on two popular benchmarks of lane detection with faster speed, which shows a superior performance-efficiency trade-off and great potential of our global association formulation.
\end{itemize}

\section{Related Works}
\subsection{Lane Detection Methods}
Lane detection aims at obtaining the accurate shape of lane lines as well as distinguishing between them.
According to the way of lane modeling, current deep-learning-based methods can be roughly divided into several categories. We will elaborate on these methods separately in this section.

\textbf{Segmentation-based methods}. Segmentation-based methods model lane line detection as a per-pixel classification problem, with each pixel classified as either lane area or background~\cite{pan2018spatial,Neven2018,Hou2019,jung2020towards}.
To distinguish different lane lines, SCNN~\cite{pan2018spatial} treats different lane lines as different categories and thus lane detection is transformed into a multi-class segmentation task. A slice-by-slice CNN structure is also proposed to enable message passing across rows and columns. 
In order to meet the real-time requirement in practice, ENet-SAD\cite{Hou2019} applies a self-attention distillation mechanism for contextual aggregation so as to allow the use of a lightweight backbone.
LaneNet~\cite{Neven2018} adopts a different way of lane representation by casting lane detection as an instance segmentation problem. A binary segmentation branch and an embedding branch are included to disentangle the segmented results into lane instances. 
Different from LaneNet, our method use offsets instead of embedding features to cluster each lane lines, which is more efficient and time-saving.

\textbf{Detection-based methods}. This kind of method usually follows a top-down manner to predict lane lines. 
Among them, anchor-based methods~\cite{linecnn,Tabelini_2021_CVPR,Xu2020} design line-like anchors and regress the offsets between sampled points and predefined anchor points. Non-Maximum Suppression (NMS) is then applied to select lane lines with the highest confidence.
LineCNN~\cite{linecnn} uses straight rays emitted from the image boundaries with certain orientations as a group of anchors. 
Curve-NAS~\cite{Xu2020} defines anchors as vertical lines and further adopts neural architecture search (NAS) to search for better backbones.
LaneATT~\cite{Tabelini_2021_CVPR} proposes an anchor-based pooling method and an attention mechanism to aggregate more global information. 
Another kind of methods~\cite{qin2020ultra, Liu_2021_ICCV} formulates lane detection as a row-wise classification problem. For each row, the model predicts the locations that possibly contain lane lines.

\textbf{Keypoint-based methods}.
Inspired by human pose estimation, some works treat lane detection as a keypoint estimation and association problem.
PINet~\cite{Ko2021} uses a stacked hourglass network~\cite{Newell2016} to predict keypoint positions and feature embedding. Different lane instances are clustered based on the similarity between feature embeddings.
FOLOLane~\cite{Qu_2021_CVPR} produces a pixel-wise heatmap with the same resolution as input to obtain points on lanes. A local geometry construction manner is also developed to associate keypoints belonging to the same lane instance.
Our GANet adopts a more efficient postprocessing approach, which needs neither feature embeddings nor local association to cluster or reconstruct the whole lane. Each keypoint finds its corresponding lane by adding its coordinate with the offset to the lane line start points in a parallel manner.

\subsection{Deformable Modeling}
Traditional CNNs are inherently limited to model irregular structures due to the fixed grid-like sampling ranges of convolution operations.
To overcome this limitation, Dai~\textit{et al}.~\cite{dai2017deformable} proposes deformable convolution to adaptively aggregate information within local areas. Compared with standard convolutions, 2D offsets obtained by an extra convolution are added at each spatial location during sampling to enable free-form deformation of the sampling grid. 
Through the learned offsets, the receptive field and the sampling location of convolutions are adaptively adjusted according to the random scale and shape of objects.
The spirit of deformable modeling has been applied in many tasks such as object detection~\cite{yang2019reppoints,deformabledetr}, object tracking~\cite{yu2020deformable} and video comprehension~\cite{xu2019learning,chen2019temporal,ying2020deformable}.
RepPoints\cite{yang2019reppoints} models an object as a set of points and predicts the offsets of these points to the object center with deformable convolutions. 
This deformable object representation provides accurate geometric localization for object detection as well as adaptive semantic feature extraction.
Ying~\textit{et al}.~\cite{ying2020deformable} proposes deformable 3D convolution to explore spatio-temporal information and realize adaptive motion comprehension for video super-resolution.
Different from these methods, our LFA module adapts to the long structure of lane lines and restricts the range of feature aggregation to adjacent points on each lane with lane-aware deformable convolutions. 

\section{Method}
\label{sec:method}
\begin{figure*}[!t]
    \centering
    \includegraphics[width=\linewidth]{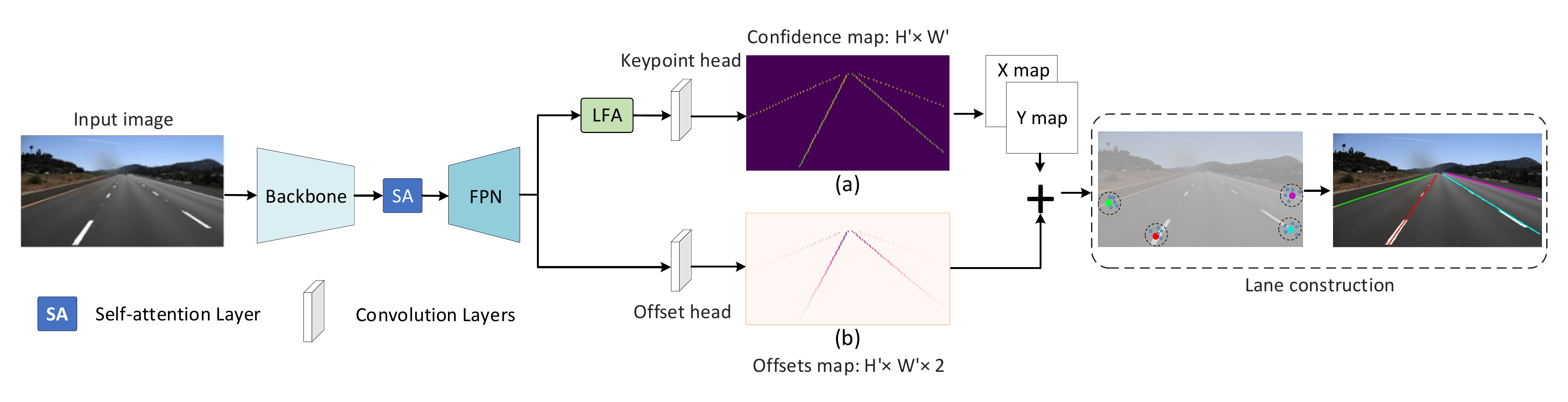}
    \caption{The overall architecture of GANet. Given a front-viewed image as input, a CNN backbone followed by a Self-Attention layer (SA) and an FPN neck are used to extract multi-scale visual features. 
    In the decoder, a keypoint head and an offset head are used to generate confidence map and offset map respectively, which are then combined to cluster keypoints into several groups, with each group indicating a lane line instance.
    Our LFA module is applied before the keypoint head to better capture local context over lane lines for keypoint estimation.}
    \label{fig:model}
\end{figure*}

The overall architecture of our proposed Global Association Network (GANet) is illustrated in Figure~\ref{fig:model}. 
Given a front-viewed image as input, a CNN backbone together with an FPN~\cite{lin2017feature} neck is adopted to extract multi-level visual representations of the input images. 
For better feature learning, a self-attention layer~\cite{vaswani2017attention} is further inserted between the backbone and the neck to obtain rich context information.
In the decoder, a keypoint head and an offset head are exploited to generate confidence map and offset map respectively. Both heads are composed of fully convolutional layers. We further devise a Lane-aware Feature Aggregator module before keypoint head to enhance the local correlations between adjacent keypoints, which facilitates to produce continuous lane lines.
For each lane instance, we first obtain its starting point as cluster centroid by selecting points with value less than $1$ over the offset map.
Afterward, keypoints belonging to the same lane are clustered around the sampled starting point with the combination of the confidence map and offset map to construct the complete lane line.

\subsection{Global Keypoint Association}
\subsubsection{Keypoint Estimation}

Given an input image $I \in \mathbb{R}^{H\times W \times 3}$, the goal of our GANet is to predict a collection of lanes $L=\{l_1, l_2, ..., l_N\}$, where $N$ is the total number of lanes, with each lane line being denoted with $K$ sampled keypoints as:
\begin{equation}
    l_i = \{p_i^1, p_i^2, ..., p_i^K\}_{i=1}^N,
    \label{equation:keypoints}
\end{equation}
where $p_i^j=(x_i^j, y_i^j)$ denotes the coordinate of the $j$-th keypoint on the $i$-th lane.
To estimate all the keypoints, we develop a keypoint head to produce a confidence map $\hat{Y} \in \mathbb{R}^{\frac{H}{r} \times \frac{W}{r}}$, where $r$ is the output stride. 
The confidence map represents the probability of each location being a keypoint on the lane.
As shown in Figure~\ref{fig:model}(a), the brighter location indicates a higher probability.

During the training phase, we sample $K$ keypoints on each lane line as ground truth keypoints and then splat them all onto a confidence map $Y \in \mathbb{R}^{\frac{H}{r} \times \frac{W}{r}}$ using a non-normalized Gaussian kernel $Y_{yx} = exp(- \frac{(x-\tilde{x})^2 + (y-\tilde{y})^2}{2 \sigma^2})$, where $\tilde{x}$ and $\tilde{y}$ denote the coordinate of each keypoint and the standard deviation $\sigma$ depends on the scale of input. 
If there is overlap between two Gaussian maps, we take the element-wise maximum between them.

We adopt penalty-reduced focal loss~\cite{lin2017focal} to deal with the imbalance between keypoint regions and non-keypoint regions as follows:

\begin{scriptsize}
\begin{equation}
\label{equation:point}
\mathcal{L}_{point}=\frac{-1}{H' \times W'} \sum_{yx} 
\left\{
    \begin{array}{lr}
         (1-\hat{Y}_{yx})^\alpha log(\hat{Y}_{yx})) & Y_{yx}=1 \\
         (1-Y_{yx})^\beta \hat{Y}_{yx}^\alpha log(1-\hat{Y}_{yx})) & otherwise,
    \end{array}
\right.
\end{equation}
\end{scriptsize}
where $\alpha$ and $\beta$ are hyper-parameters of focal loss and $H' \times W'$ denotes $\frac{H}{r} \times \frac{W}{r}$. The subscript $yx$ represents obtaining the value at coordinate $(x,y)$. 

Due to the output stride $r$, the point $(x_i^j, y_i^j)$ of the input image is mapped to the location $(\lfloor \frac{x_i^j}{r} \rfloor, \lfloor \frac{y_i^j}{r} \rfloor)$, which can cause performance degradation. 
To address this quantization error, we additionally predict a compensation map $\hat{\delta}_{yx}$ and apply L1 loss to keypoint locations only:
\begin{equation}
\label{equation:quant}
        \mathcal{L}_{quant}= \frac{1}{H' \times W'} \sum_{yx}{\left|\hat{\delta}_{yx} - \delta_{yx} \right|},
\end{equation}
where $ \delta_{yx} = (\frac{x_i^j}{r} - \lfloor \frac{x_i^j}{r} \rfloor, \frac{y_i^j}{r} - \lfloor \frac{y_i^j}{r} \rfloor)$ denotes the ground truth of quantization compensation map.
This part is not shown in Figure~\ref{fig:model} for simplicity.

\subsubsection{Starting Point Regression}
\label{sec:regress}
To distinguish different lane lines, we propose to use the starting point to represent each lane instance uniquely due to its stability and largest margins between each other.
Instead of regressing the absolute coordinate $(sx_i, sy_i)$ of the starting point directly, we regress the offset from each keypoint to it, which can be defined as:
\begin{equation}
\label{equation: displacement}
    (\Delta x_i^j, \Delta y_i^j) = (sx_i, sy_i) - (x_i^j, y_i^j),
\end{equation}

Thus, we can generate the ground truth offset map $O_{yx}$ with the shape of $\frac{H}{r} \times \frac{W}{r} \times C$. In particular, the subscript $yx$ denotes the value on location $(x_i^j, y_i^j)$ which is equal to $(\Delta x_i^j, \Delta y_i^j)$ while other locations have zero values. $C=2$ contains the x-direction and y-direction offsets respectively. 

In order to estimate the offset map $\hat{O}_{yx}$, we introduce an offset head as shown in Figure~\ref{fig:model}. Similarly, L1 loss is used to constrain the offset map as follows:
\begin{equation}
\label{equation:offsetloss}
    \mathcal{L}_{offset}=\frac{1}{H' \times W'} \sum_{yx}{\left| \hat{O}_{yx} - O_{yx} \right|},
\end{equation}
The supervision applies only on keypoint locations and the rest locations are ignored.

\begin{figure}[!t]
    \centering
    \includegraphics[scale=0.35]{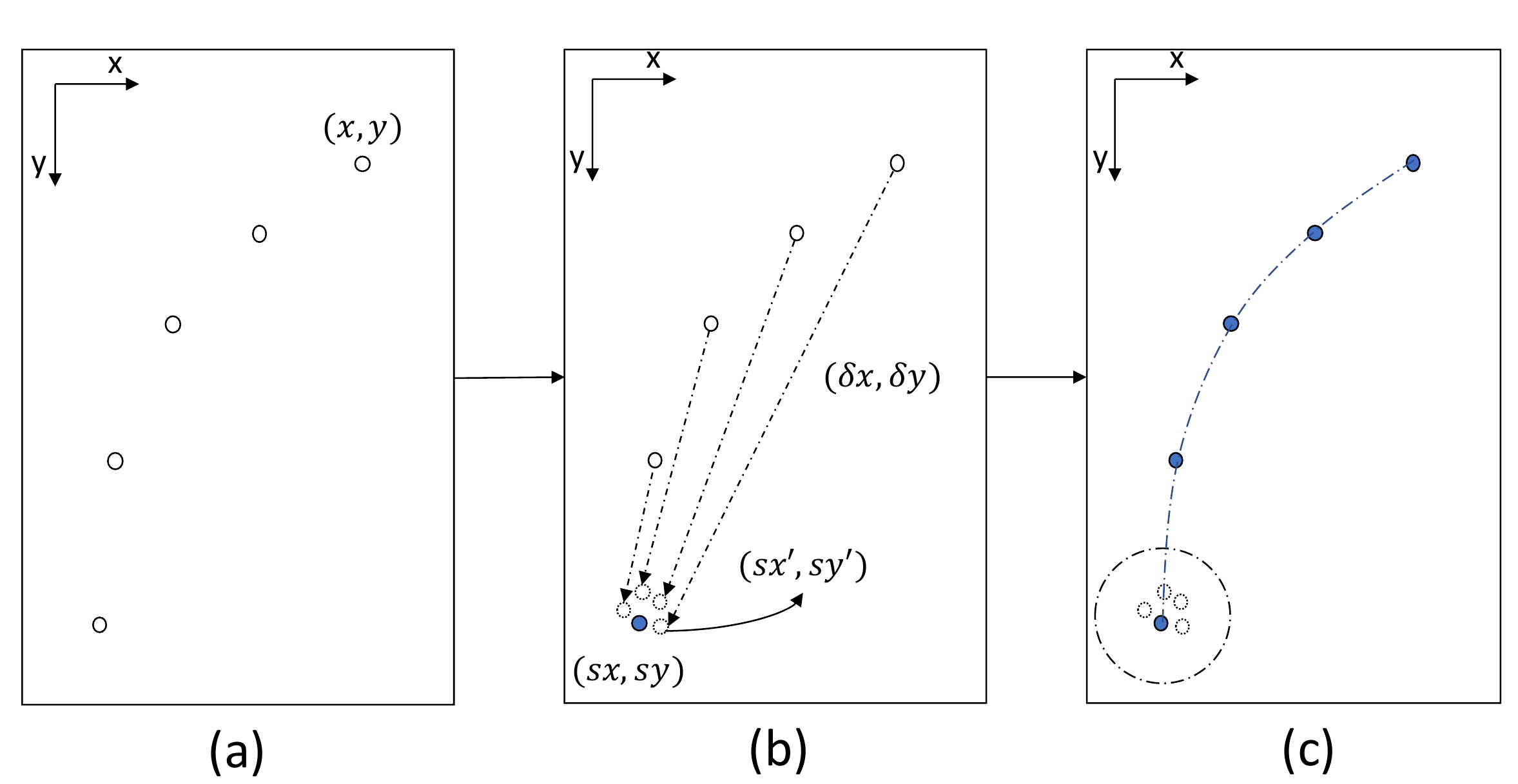}
    \caption{Illustration of lane construction. (a) Valid keypoints are selected from the confidence map. $(x,y)$ is taken as an example. (b) Starting point $(sx, sy)$ (blue dot) is sampled first. The rest keypoints point to the starting point with the predicted offset $(\delta x, \delta y)$ and estimate the coordinate of the starting points as $(sx', sy')=(x,y)+(\delta x, \delta y)$ (hollow dots). (c) Keypoints that point to the neighbourhood of starting point $(sx, sy)$ are grouped as a whole lane.}
    \label{fig:postprocess}
\end{figure}

\subsubsection{Lane Construction}

The pipeline of lane construction is presented in Figure~\ref{fig:postprocess}, which includes obtaining the locations of all the possible lane points and then grouping them into different lane instances. 
We first apply a $1\times 3$ max pooling layer on the keypoint confidence map $\hat{Y}$ to select points of maximum responses within a horizontal local region as valid keypoints, which is shown in Figure~\ref{fig:postprocess}(a).
Then, we group them to describe each lane as an ordered list of keypoints as follows:
\begin{equation}\label{equa root}
    l = \{(sx, sy), (x^2, y^2), (x^3, y^3), ..., (x^K, y^K)\},
\end{equation}
where $(sx, sy)$ denotes the starting point of the lane and $(x^j, y^j), j\in [2, K]$ are the subsequent keypoints. 

To obtain the starting point of each lane, we select keypoints whose values are less than $1$ on the offset map as the candidate starting points.
Since there might be multiple keypoints matching the above criteria within the same local region, the geometric center point of the region is chosen to ensure uniqueness.
By this means, instances of all the lanes are preliminarily determined with their starting points.

Afterward, we associate the rest keypoints to their belonged lanes according to the estimated offsets between keypoints and corresponding starting points, which is shown in Figure~\ref{fig:postprocess}(b).
Each keypoint estimates the coordinate of the lane line starting point as follows:
\begin{equation}
\label{equation:startpoing}
(sx', sy') = (x, y) + (\delta x, \delta y),
\end{equation}
where $(x, y)$ is the coordinate of the observed keypoint and $(\delta x, \delta y) = O_{yx}$ denotes the corresponding offset obtained in Section~\ref{sec:regress}.
The keypoint $(x,y)$ is associated to the $i$-th lane only if the distance between $(sx', sy')$ and $(sx, sy)$ is less than a predefined threshold $\theta_{dis}$.
As shown in Figure~\ref{fig:postprocess}(c), keypoints that point to the neighborhood of the same starting point are grouped to produce a whole lane. The above procedures are done by matrix operations to ensure parallel keypoints association.

\subsection{Lane-aware Feature Aggregator}
\begin{figure}[!t]
    \centering
    \includegraphics[scale=0.3]{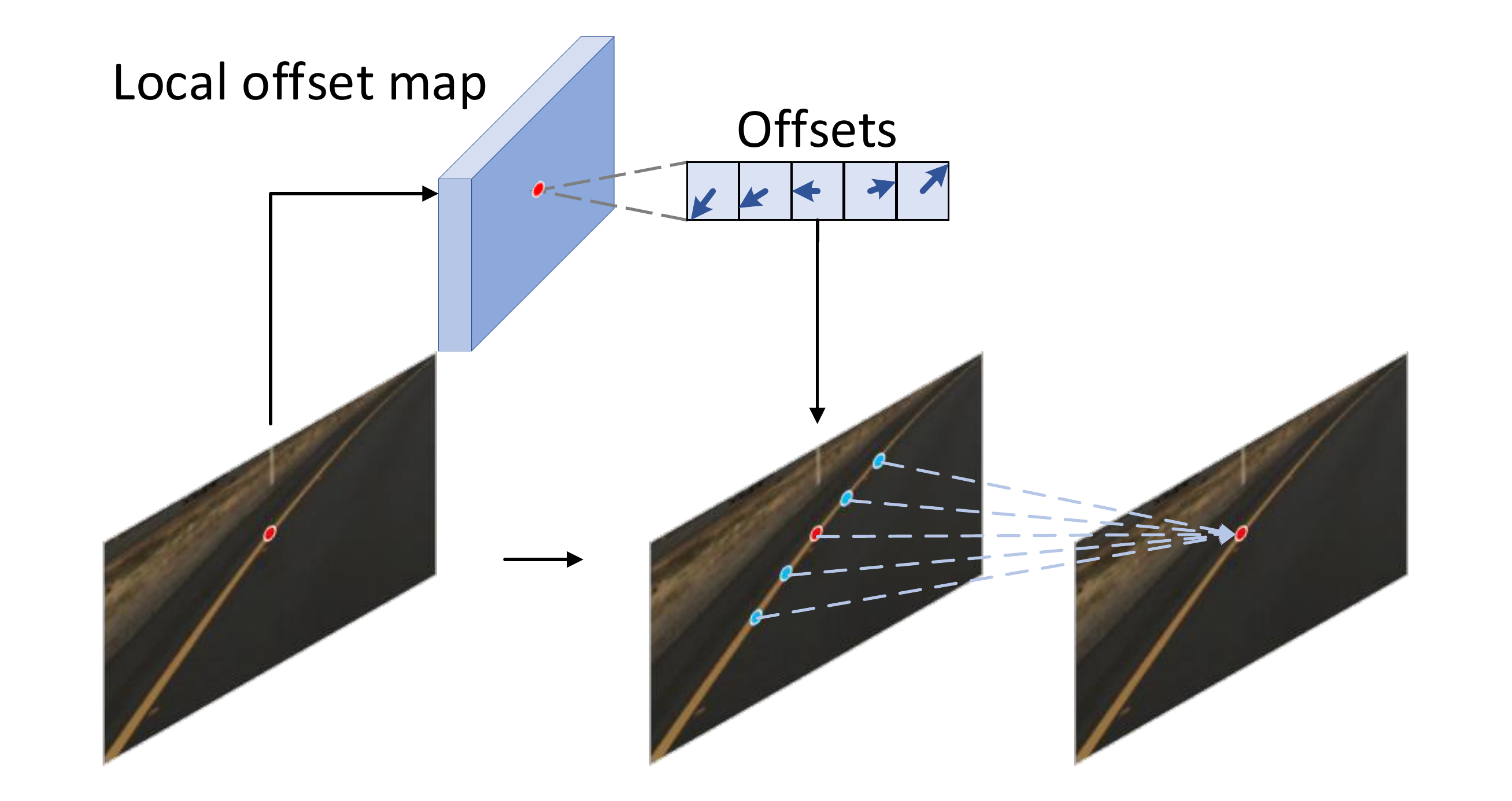}
    \caption{Illustration of LFA module. The red dot denotes the observed keypoint. We first predict offsets between the red dot and its adjacent keypoints (in blue) and then gather features of these keypoints to enhance the context of the red keypoint.}
    \label{fig:dconv}
\end{figure}

Traditional 2D convolutions sample features within a fixed grid-like region, which is not suitable for handling the slender shapes of lane lines. 
Inspired by Dai~\textit{et al}.~\cite{dai2017deformable}, we propose a Lane-aware Feature Aggregator (LFA) module to adaptively gather information from adjacent points on the lanes, so as to enhance the local feature representation of each keypoint.
The illustration of our LFA module is shown in Figure~\ref{fig:dconv}. 
Take a specific keypoint as an example, we first use a convolution layer to predict the offset between it and its surrounded $M$ keypoints on the same lane as follows:
\begin{equation}
    \Delta{P_i}=\phi(\mathcal{F}(p_i)), 
    \label{equation:offset}
\end{equation}
where $p_i$ denotes the coordinate of the $i$-th keypoint, $\mathcal{F}(p_i)$ denotes the feature representation of the $i$-th keypoint and $\Delta{P_i} = \{\Delta{p_i^m}|m=1,...,M\} \in \mathbb{R}^{2M}$ denotes the predicted offsets. 
Afterwards, features of adjacent points are integrated with a deformable convolution to aggregate context of the $i$-th keypoint as:
\begin{equation}
    \hat{\mathcal{F}}(p_i)=\sum_{m=1}^M{w_m\cdot\mathcal{F}(p_i+\Delta{p_i^m})},
    \label{equation:sumup}
\end{equation}
where $w_m, m=1,...,M$ is the weight of the convolution and $(\cdot)$ means multiplication. 

To enhance the ability of LFA for learning the local shapes of lane lines, we further introduce an auxiliary loss to supervise the offsets $\Delta{P_i}$.
We denote the ground truth of offsets between the $i$-th keypoint and the keypoints on the corresponding lane line as $\Delta{G_i}=\{\Delta{g_i^k}|k=1,...,K\}$, which is calculated with $\Delta{g_i^k}={g_i^k-p_i}$, where $g_i^k$ is the ground-truth coordinate of the $k$-th keypoint on the same lane line with the $i$-th keypoint.

\begin{figure}[!t]
    \centering
    \includegraphics[scale=0.45]{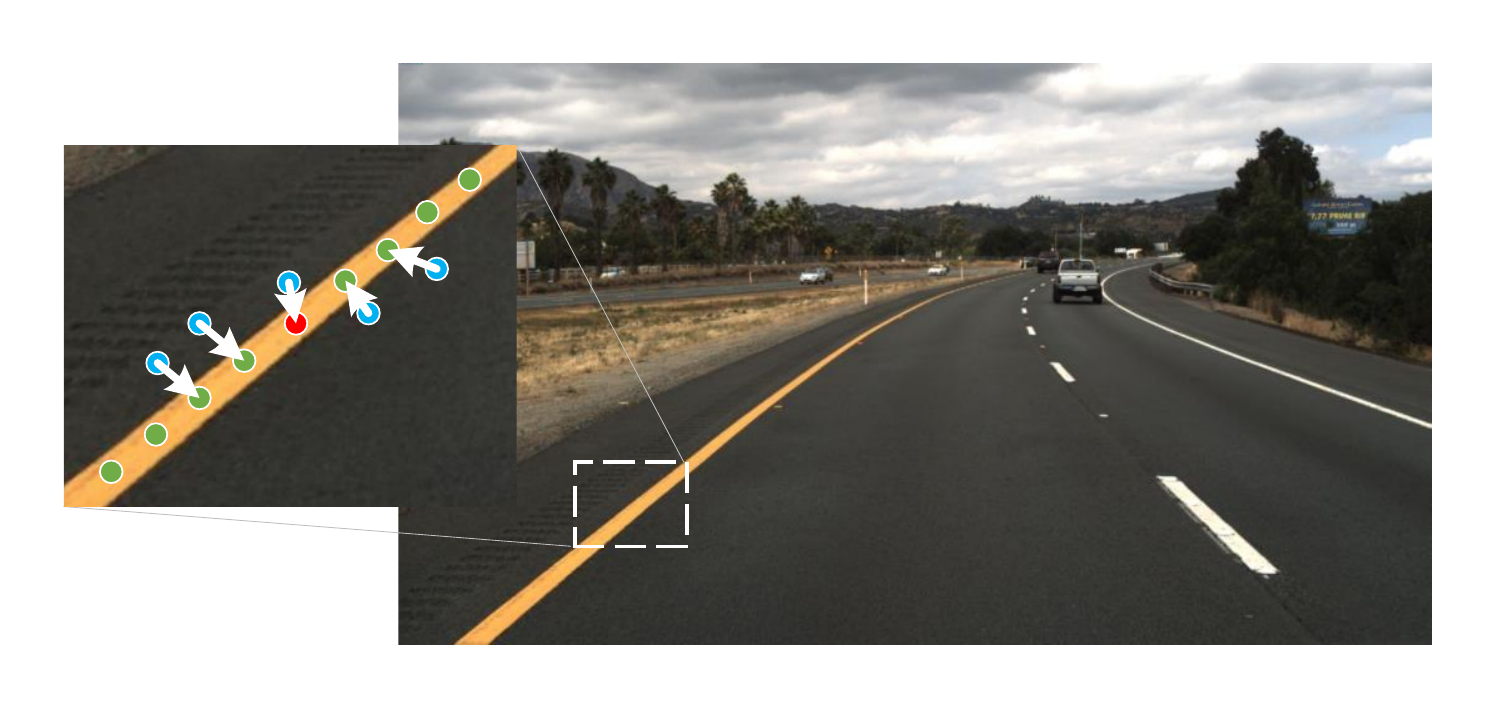}
    \caption{Illustration of the matching between predict points and their ground truth. The red dot is the observed keypoint. The blue dots are the predicted locations of adjacent keypoints. The green dots are the ground-truth locations of adjacent keypoints on the lane line.}
    \label{fig:match}
\end{figure}

As is shown in Figure~\ref{fig:match}, a matching need to be established between $\Delta{p_i}$ and $\Delta{g_i}$. We search for an assignment $\sigma$ with the lowest matching cost:
\begin{equation}
    \hat{\sigma}=\mathop{\arg\min}_{\sigma}{\sum_m^{M}\mathcal{L}_{match}(\Delta{p_i^m}, \Delta{{g_i}}^{\sigma(m)})},
\end{equation}
where $\mathcal{L}_{match}=L_2(\Delta{p_i^m}, \Delta{{g_i}}^{\sigma(m)})$.
Following prior works~\cite{stewart2016end,carion2020end}, the Hungarian algorithm is adopted to efficiently compute the optimal assignment. SmoothL1 loss is then applied to supervise the prediction of adjacent keypoints:
\begin{equation}
    \mathcal{L}_{aux}=\frac{1}{KNM}\sum_{i=1}^{KN}\sum_{m=1}^M{SmoothL1(\Delta{p_i^m}, \Delta{{g_i}}^{\hat{\sigma}(m)})},
\end{equation}
where $K$ denotes the number of keypoints on each lane line, $N$ denotes the number of lane lines and $M$ denotes the number of sampled adjacent keypoints.

The total loss function is the combination of different losses with corresponding coefficients:
\begin{equation}
\label{equation:total loss}
\begin{split}
\mathcal{L}_{total} = \lambda_{point} \mathcal{L}_{point} + \lambda_{quant} \mathcal{L}_{quant} \\  + \lambda_{offset} \mathcal{L}_{offset} + \lambda_{aux} \mathcal{L}_{aux}.
\end{split}
\end{equation}

\section{Experiments}

In this section, we first introduce the experimental setting of our method. The next subsection discusses the results for each dataset. Ablation experiments for each module is presented in the last subsection.
\subsection{Experimental Setting}
\subsubsection{Datasets and Evaluation Metrics}
We conduct experiments on two popular lane detection benchmarks including CULane~\cite{pan2018spatial} and TuSimple~\cite{Tusimple}.

\textbf{CULane:} CULane dataset contains $88,880$ training images and $34,680$ testing images, including both urban and highway scenes. The test images are classified as $9$ different scenarios. F1 measure is the only metric for evaluation, which is based on IoU. A predicted lane whose IoU is greater than $0.5$ is judged as true positive (TP), otherwise false positive (FP) or false negative (FN). F1 measure is defined as the harmonic average of precision and recall.

\textbf{TuSimple:} TuSimple is a real highway dataset which consists of $3,626$ images for training and $2,782$ images for testing. The main evaluation metric of TuSimple dataset is accuracy, which is formulated as follows:
$$accuracy=\frac{\sum_{clip}{C_{clip}}}{\sum_{clip}{S_{clip}}}$$
where $C_{clip}$ is the number of points correctly predicted by the model and $S_{clip}$ is the total number of points in the clip (or image). A predicted point is considered correct only if it is within $20$ pixels to the ground truth point. The predicted lane with accuracy greater than $85\%$ is considered as a true positive. We also report the F1 score in the following experiments.

\begin{table}[!htbp]
    \centering
    \scalebox{0.85}{
        \begin{tabular}{lccc}
            \hline
            Model version & Backbone & FPN layers & Output downscale \\ \hline
            GANet-S & resnet-18  & 3 & 8 \\
            GANet-M & resnet-34  & 3 & 8 \\
            GANet-L & resnet-101 & 4 & 4 \\
            \hline
        \end{tabular}
    }
    \caption{Details of different versions of GANet.}
    \label{tab:version}
\end{table}

\begin{table*}[t]
    \begin{center}
    \scalebox{0.90}{
        \begin{tabular}{lcccccccccccc}
            \hline
            Method& Total & Normal & Crowded & Dazzle & Shadow & No line & Arrow & Curve & Cross & Night & FPS \\ \hline
            \textbf{Segmentation-based} \\ \hline
            SCNN~\cite{pan2018spatial} & 71.60 & 90.60 & 69.70 & 58.50 & 66.90 & 43.40 & 84.10 & 64.40 & 1990 & 66.10 & 7.5 \\
            ENet-SAD~\cite{hou2019learning} & 70.80 & 90.10 & 68.80 & 60.20 & 65.90 & 41.60 & 84.00 & 65.70 & 1998 & 66.00 & 75 \\    \hline
            \textbf{Detection-based} \\ \hline
            FastDraw~\cite{philion2019fastdraw} & \text{-} & 85.90 & 63.60 & 57.00 & 69.90 & 40.60 & 79.40 & 65.20 & 7013 & 57.80 & 90.3 \\
            UFAST-ResNet18~\cite{qin2020ultra} & 68.40 & 87.70 & 66.00 & 58.40 & 62.80 & 40.20 & 81.00 & 57.90 & 1743 & 62.10 & \textbf{322.5} \\
            UFAST-ResNet34~\cite{qin2020ultra} & 72.30 & 90.07 & 70.20 & 59.50 & 69.30 & 44.40 & 85.70 & 69.50 & 2037 & 66.70 & 175 \\
            ERF-E2E~\cite{yoo2020end} & 74.00 & 91.00 & 73.10 & 64.50 & 74.10 & 46.60 & 85.80 & 71.90 & 2022 & 67.90 & \text{-}\\
            CurveLanes-NAS-L~\cite{Xu2020} & 74.80 & 90.70 & 72.30 & 67.70 & 70.10 & 49.40 & 85.80 & 68.40 & 1746 & 68.90 & \text{-} \\
            LaneATT-ResNet18~\cite{Tabelini_2021_CVPR} & 75.13 & 91.17 & 72.71 & 65.82 & 68.03 & 49.13 & 87.82 & 63.75 & \textbf{1020} & 68.58 & 250 \\
            LaneATT-ResNet34~\cite{Tabelini_2021_CVPR} & 76.68 & 92.14 & 75.03 & 66.47 & 78.15 & 49.39 & 88.38 & 67.72 & 1330 & 70.72 & 171 \\
            LaneATT-ResNet122~\cite{Tabelini_2021_CVPR} & 77.02 & 91.74 & 76.16 & 69.47 & 76.31 & 50.46 & 86.29 & 64.05 & 1264 & 70.81 & 26 \\    \hline
            \textbf{Keypoint-based} \\ \hline
            FOLOLane-ERF~\cite{Qu_2021_CVPR} & 78.80 & 92.70 & 77.80 & \textbf{75.20} & 79.30 & 52.10 & 89.00 & 69.40 & 1569 & \textbf{74.50} & 40 \\  
            \rowcolor{mygray}
            GANet-S           & 78.79 & 93.24 & 77.16 & 71.24 & 77.88 & \textbf{53.59}  & 89.62 & 75.92 & 1240 & 72.75 & 153 \\
            \rowcolor{mygray}
            GANet-M           & 79.39 & \textbf{93.73} & 77.92 & 71.64 &  \textbf{79.49} & 52.63 & \textbf{90.37} & 76.32  &  1368 & 73.67 & 127 \\
            \rowcolor{mygray}
            GANet-L           & \textbf{79.63} &  93.67 & \textbf{78.66} & 71.82 & 78.32 & 53.38  & 89.86 & \textbf{77.37} & 1352 & 73.85 & 63 \\  \hline
        \end{tabular}
    }
    \caption{Comparison with state-of-the-art methods on CULane test set. The evaluation metric is F1 score with IoU threshold=0.5. For cross scenario, only FP are shown.}
    \label{tab:culane}
    \end{center}
\end{table*}

\subsubsection{Implementation Details}
We choose ResNet-18, ResNet-34 and ResNet-101~\cite{He_2016_CVPR} as the backbones to form three different versions of GANet, which are referred as GANet-S, GANet-M and GANet-L. The detail of each version is shown in Table~\ref{tab:version}.
We first resize the input images to $800\times 320$ during the training and testing phases. The number of sampled points in LFA is set as $M=9$. The loss weights are set as $\lambda_{point}=1.0$, $\lambda_{quant}=1.0$, $\lambda_{offset}=0.5$, $\lambda_{aux}=1.0$. 
The hyper-parameters $\alpha$ and $\beta$ in Equation~\ref{equation:point} are set as $2$ and $4$ respectively. 
For optimization, we used Adam optimizer and poly learning rate decay with an initial learning rate of $0.001$. We train $300$ and $40$ epochs for Tusimple and CULane respectively with a batchsize of $32$ per GPU. Data augmentation is applied to the training phase, including random scaling, cropping, horizontal flipping, random rotation, and color jittering. In the test phase, we set the threshold of keypoints as $0.4$ and $\theta_{dis}$ for keypoint association as $4$. Training and testing are both performed on Tesla-V100 GPUs.

\begin{table}[!htbp]
    \scalebox{0.82}{
        \begin{tabular}{lccccc}
            \hline
            \textbf{Method} & F1 & Acc & FP & FN & FPS \\ \hline
            \textbf{Segmentation-based} \\ \hline
            SCNN~\cite{pan2018spatial} & 95.97 & 96.53 & 6.17 & \textbf{1.80}  & 7.5 \\
            EL-GAN~\cite{ghafoorian2018gan} & 96.26 & 94.90 & 4.12 & 3.36  & 10 \\
            ENet-SAD~\cite{hou2019learning} & 95.92 & 96.64 & 6.02 & 2.05 & 75 \\ \hline
            \textbf{Detection-based} \\ \hline
            FastDraw~\cite{philion2019fastdraw} & 93.92 & 95.20 & 7.60 & 4.50 & 90.3  \\
            UFAST-ResNet18~\cite{qin2020ultra} & 87.87 & 95.82 & 19.05 & 3.92 & 312.5 \\
            UFAST-ResNet34~\cite{qin2020ultra} & 88.02 & 95.86 & 18.91 & 3.75 & 169.5  \\
            ERF-E2E~\cite{yoo2020end} & 96.25 & 96.02 & 3.21 & 4.28 & \text{-} \\ 
            LineCNN~\cite{li2019line} & 96.79 & 96.87 & 4.42 & 1.97 & 30 \\
            LaneATT-ResNet18~\cite{Tabelini_2021_CVPR} & 96.71 & 95.57 & 3.56 & 3.01  & 250  \\
            LaneATT-ResNet34~\cite{Tabelini_2021_CVPR} & 96.77 & 95.63 & 3.53 & 2.92  & 171  \\
            LaneATT-ResNet122~\cite{Tabelini_2021_CVPR} & 96.06 & 96.10 & 5.64 & 2.17  & 26  \\ \hline
            \textbf{Other Methods} \\ \hline
            PolyLaneNet~\cite{tabelini2020polylanenet} & 90.62 & 93.36 & 9.42 & 9.33  & 115 \\
            LSTR~\cite{liu2021end} & 96.86  & 96.18 & 2.91 & 3.38  & \textbf{420} \\ \hline
            \textbf{Keypoint-based} \\ \hline
            FOLOLane-ERF~\cite{Qu_2021_CVPR}    & \text{-}  & \textbf{96.92} & 4.47 & 2.28 & 40 \\
            \rowcolor{mygray}
            GANet-S           & \textbf{97.71} & 95.95 & \textbf{1.97} & 2.62 & 153 \\
            \rowcolor{mygray}
            GANet-M           & 97.68 & 95.87 & 1.99 & 2.64 & 127 \\
            \rowcolor{mygray}
            GANet-L           & 97.45 & 96.44 & 2.63 & 2.47 & 63 \\ \hline
        \end{tabular}
    }
    \caption{Comparison with state-of-the-art methods on TuSimple test set.}
    \label{tab:tusimple}
\end{table}

\subsection{Quantitative Results}

\subsubsection{Results on CULane}
The results on CULane test set are shown in Table~\ref{tab:culane}. Our GANet-L achieves the state-of-the-art result on CULane dataset with $79.63\%$ F1 score and $63$ FPS, which exceeds models of similar size, like LaneATT-ResNet122, with large margins on both performance and speed.
Compared with another keypoint-based method, FOLOLane-ERF~\cite{Qu_2021_CVPR}, our GANet-S achieves a comparable performance of $78.79\%$ F1 score but runs $3.8$ times faster, which shows a superior trade-off between performance and efficiency and demonstrate the speed advantage of our global association formulation.
Furthermore, our methods achieve the highest F1 score in six scenarios, especially in Curve scenario.
Our GANet-L achieves $77.37\%$ in this scenario and outperforms previous state-of-the-art method, ERF-E2E~\cite{yoo2020end}, with more than $5\%$, indicating the superiority of our method in describing complex lane line shapes.

\subsubsection{Results on TuSimple}

The comparison results on TuSimple test set are shown in Table~\ref{tab:tusimple}. Our GANet-S outperforms all other methods and achieves the highest F1 score of $97.71\%$ with high FPS. 
It is worth noting GANet-S exceeds UFast-ResNet34 and LaneATT-ResNet34 which have similar speed with large margins, showing the great potential of our global association formulation.
Similar to LaneATT~\cite{Tabelini_2021_CVPR}, enlarging model capacity does not necessarily bring performance improvement. It is possibly because of the small amount and the single scenario of Tusimple dataset. Results have already been saturated and a larger model may cause the overfitting problem. 

\begin{figure*}[!t]
    \centering
    \includegraphics[scale=0.18]{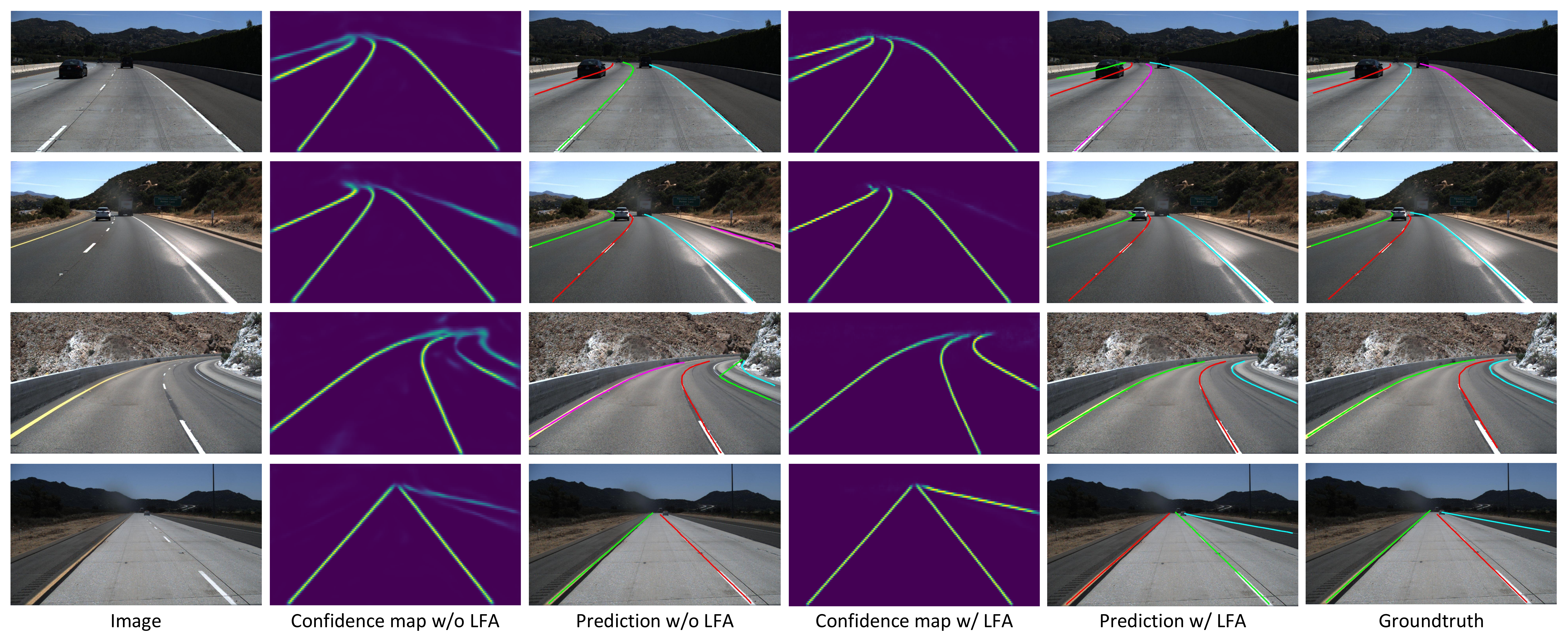}
    \caption{Visualization results of GANet w/wo LFA. The first column is the input image. The second and third columns are the predicted point confidence map and lane lines without LFA. The fourth and fifth columns are the predicted point confidence map and lane lines with LFA. The last column is the ground-truth lane lines}
    \label{fig:LFAmodule}
\end{figure*}

\subsubsection{Ablation Study}
To explore the properties of our proposed LFA module, we conduct an ablation study on the CULane dataset. All the following experiments are based on the small version of GANet. Results are shown in Table~\ref{tab:ablation}.
The first row shows the baseline method without our LFA module.
In the second row, the LFA module is integrated into GANet without auxiliary loss.
The last row shows the result of our whole GANet.

From the first two rows we can observe that LFA module without auxiliary loss is effective for lane line detection, which is due to flexible integration of context.
Comparing the last two rows, we can also find that the auxiliary loss is vital to the LFA module, which can guide LFA to focus on the key information on the lane line. The visualization analysis is performed in Section~\ref{sec:visialization}.

\begin{table}[!htbp]
    \centering
    \scalebox{1.0}{
        \begin{tabular}{cccc}
            \hline
            Baseline & LFA & AuxLoss & F1 \\ \hline
            \checkmark &  &  & 77.84 \\
            \checkmark & \checkmark &  & 78.30 \\ 
            \checkmark & \checkmark & \checkmark & 78.79 \\ \hline
        \end{tabular}
    }
    \caption{Ablation study of LFA module}
    \label{tab:ablation}
\end{table}

\subsection{Qualitative results\label{sec:visialization}}
We visualize the qualitative results w/wo LFA in Figure~\ref{fig:LFAmodule}. The $2$-nd and $4$-th columns are the visualization of confidence map without and with LFA correspondingly. 
As shown in the results from the first row, the LFA module makes correct prediction even with vehicle occlusion due to the fact that predicted lane points enhance each other.
From the results in second and third rows, it can also be concluded that the LFA module is able to suppress background noise which may be introduced by global attention.

\begin{figure}[!htbp]
    \centering
    \includegraphics[scale=0.43]{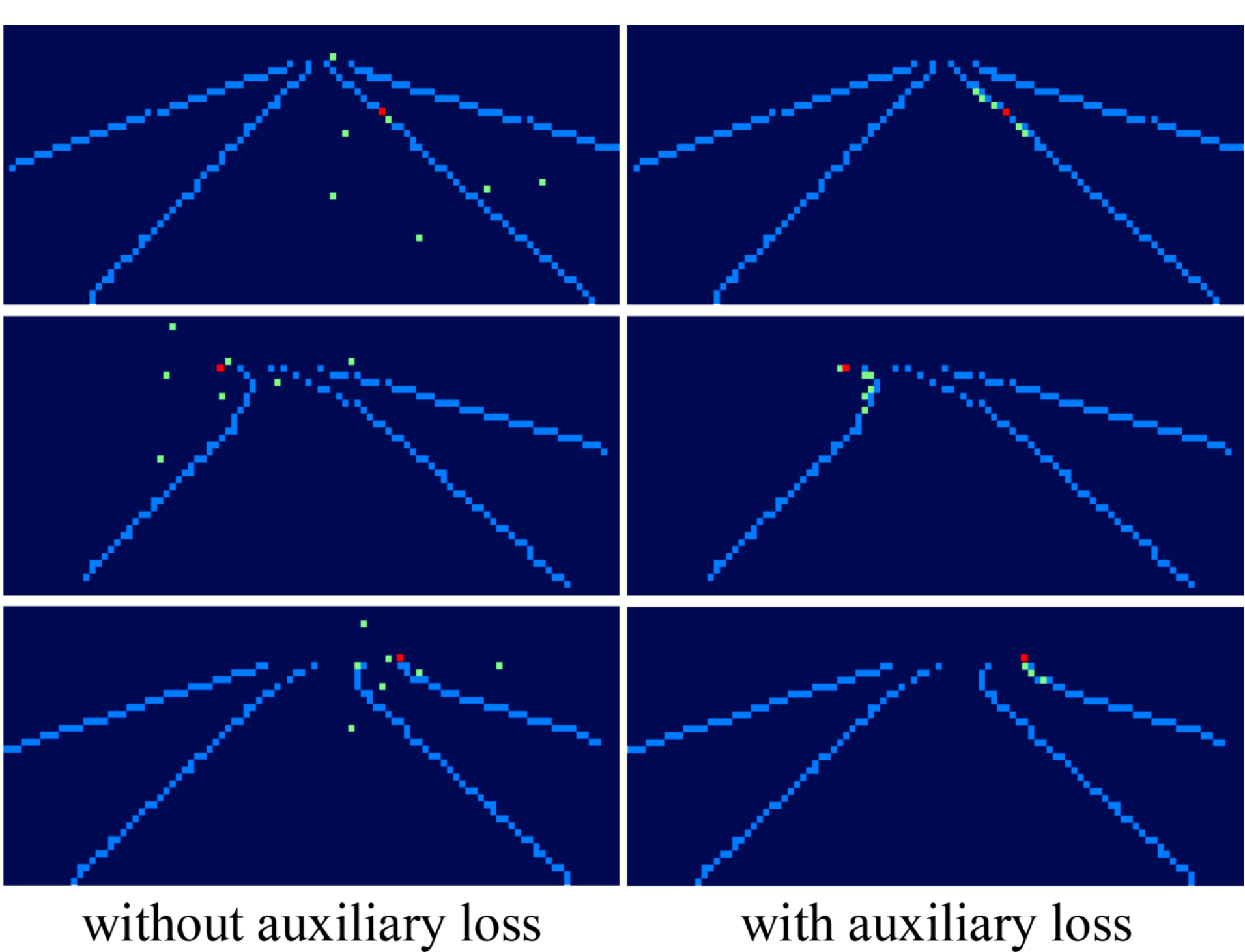}
    \caption{Visualization results of LFA w/wo auxiliary loss. The red point is the observation point. The green points are the predicted aggregation points. The light blue points are the ground-truth points on the lane line.}
    \label{fig:dpoints}
\end{figure}

To intuitively investigate the properties of the LFA module, we visualize the predicted feature aggregation points in Figure~\ref{fig:dpoints}. The first row shows a common straight lane case. With the addition of auxiliary losses, the LFA module can predict the aggregation points around the lane line. Meanwhile, the predicted aggregation points are irregular without the auxiliary loss. The last two rows show the aggregation points in the curved lane case. It is demonstrated that the LFA module is robust in its understanding of the local structures of the lane lines. This property contributes to the enhancement of lane line features and the suppression of background noise.

\section{Conclusion and Discussion}
In this paper, we propose a Global Association Network (GANet) to formulate the lane detection problem from a new perspective, where each keypoint is directly regressed to the starting point of the lane line instead of point-by-point extension.
The association of keypoints to their belonged lane line is conducted by predicting their offsets to the corresponding starting points of lanes globally, which greatly improves the effectiveness. 
We further propose a Lane-aware Feature Aggregator (LFA) to adaptively capture the local correlations between adjacent keypoints to supplement local information.
Experimental results show our GANet outperforms previous methods with higher speed.

\textbf{Limitation}. The limitation of our method is that the offsets to the starting point may become difficult to regress when the output stride is set to $1$ due to the large absolute value of the offsets. 
In the future, we hope to address this problem by regressing the offsets with multiple levels to alleviate the regression difficulty.

{\small
\bibliographystyle{ieee_fullname}
\bibliography{submission_ref}
}

\end{document}